\pdfoutput=1
\documentclass[11pt,letterpaper]{article}

\usepackage{times}
\usepackage{epsfig}
\usepackage{graphicx}
\usepackage{amsmath}
\usepackage{amssymb}
\usepackage{subfigure}
\usepackage[margin=1in]{geometry}

\usepackage[pagebackref=true,breaklinks=true,letterpaper=true,colorlinks,bookmarks=false]{hyperref}

\begin{document}

\title{Real-time Multiple People Hand Localization in 4D Point Clouds}

\author{Hao Jiang and Quanzeng You\\
Microsoft, USA
}
\date{}
\maketitle

\begin{abstract}
We propose novel real-time algorithm to localize hands and find their associations with multiple people
in the cluttered 4D volumetric data (dynamic 3D volumes). Different from the traditional
multiple view approaches, which find key points in 2D and then triangulate
to recover the 3D locations, our method directly processes the dynamic 3D data
that involve both clutter and crowd.
The volumetric representation is more desirable than the partial observations
from different view points and enables more robust and accurate results.
However, due to the large amount of data in the volumetric representation
brute force 3D schemes are slow.
In this paper, we propose novel real-time methods to tackle the problem
to achieve both higher accuracy and faster speed than previous approaches.
Our method detects the 3D bounding box of each subject and localizes the hands of each person.
We develop new 2D features for fast candidate proposals and
optimize the trajectory linking using a new
max-covering bipartite matching formulation, which is critical for robust performance.
We propose a novel decomposition method to reduce the key point localization in
each person 3D volume to a sequence of efficient 2D problems.
Our experiments show that the proposed method is
faster than different competing methods
and
it gives almost half the localization error.
\end{abstract}

\vspace{-10pt}
\section{Introduction}
\begin{figure}[tb]
\centering
\includegraphics[width=0.6\linewidth]{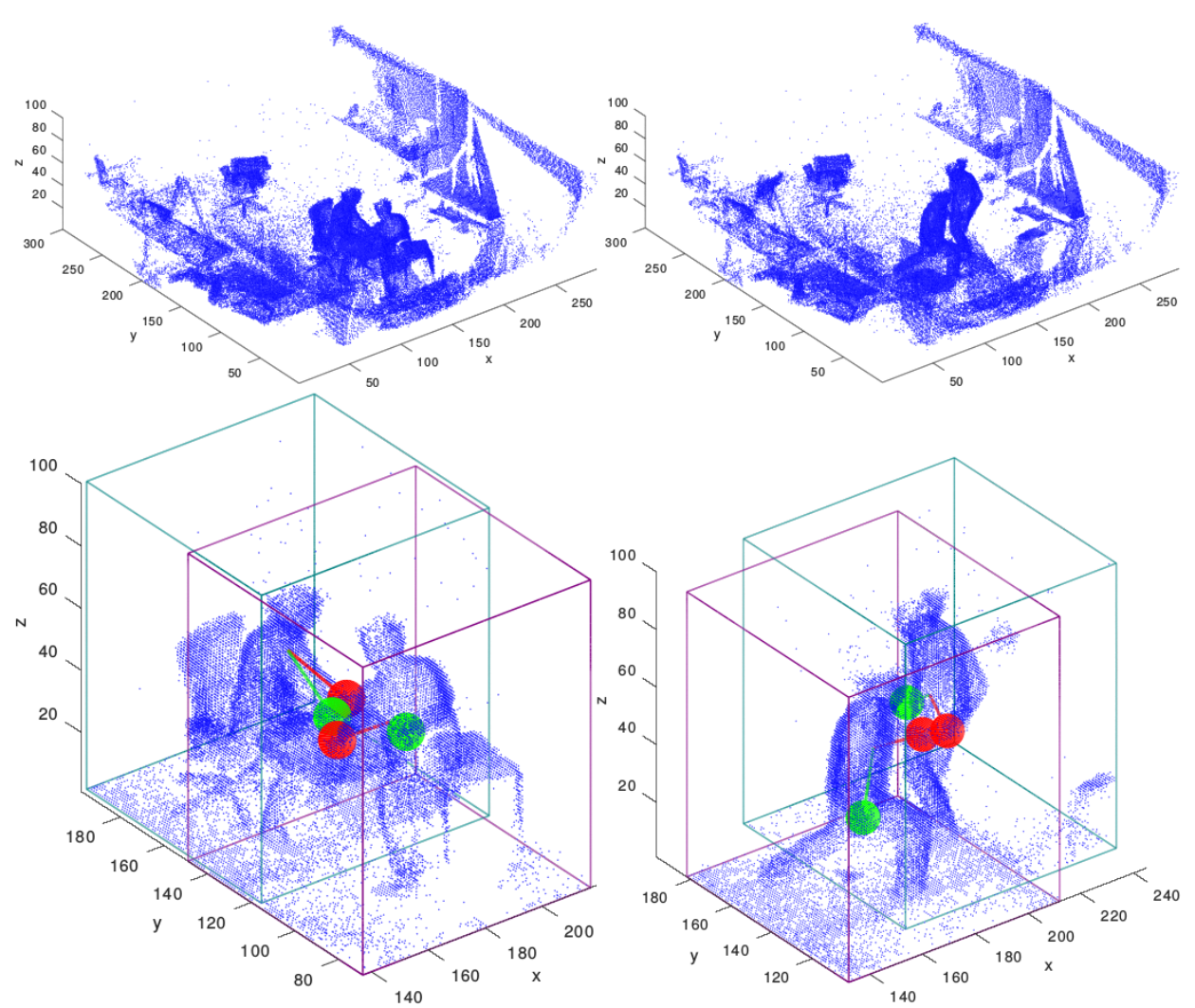}%
\caption{Finding people's hands and associations in 4D (dynamic 3D volume).
Row 1: 3D volumes at two time instants. The volumetric data is generated in real time
from seven Kinect-v2 RGBD cameras. 
Row 2: Our method tracks the 3D bounding boxes of each subject and
finds the left (red) and right (green) hands and their associations (red or green line between hands and the
volume centers) with each person.
}
\label{fig-teaser}
\vspace{-10pt}
\end{figure}

\vspace{-5pt}
Finding people's hands and their associations with each instance
in a crowded and cluttered scene is a critical computer vision task that enables many different applications,
such as human computer interaction, gesture recognition, and activity recognition.
In this paper, we propose novel real-time algorithm to tackle the problem, as illustrated
in Fig.~\ref{fig-teaser}.
The volume data can be
generated from multiple stereo
cameras, time-of-flight cameras, LiDAR or even non-camera devices.
The dynamic 3D volume is a holistic 4D representation of the scene and objects.

The strategy of our approach is that we directly process
the 4D data so that we can take advantage of a unified representation
which is more complete than images from any view points.
Most of the previous methods \cite{3dpose-survey, pose-multiview2, pose-multiview1, cmudome}
for people 3D key point localization
do not directly work on the volume representation; they process images from different camera views
and optimize the 3D key point localization using the multiple view results.
Handling self- and inter-occlusions is a challenging task for these traditional methods.


Directly working on the 4D data also presents unique challenge: the very large amount
of data that needs to be processed. The voxelized volume
becomes a much bigger data structure than the original 2D data representation.
This gives
an impression that processing such data would be very slow. However, with properly designed algorithms
we can achieve real-time performance with even better results.
The 4D volume is also often cluttered due to the sensor
noise and misalignment in time and space
among partial point clouds from different views.
It involves
complex interactions among people and between people and objects.

We propose faster and more accurate method to solve the problem.
Our method first finds each individual person in the 4D data.
We develop new 2D feature representations for fast people proposals and
optimize the data association using a new
max-covering bipartite matching formulation, which is key for robust people detection and tracking.
With each tracked person 3D volume, we localize the left and right hand of each person.
We propose a novel decomposition method to reduce the key point localization in
each person 3D volume to a sequence of 2D problems.
The decomposition scheme enables the proposed method
to achieve high efficiency
and
at the same time it gives much smaller error than the competing methods.

The contribution of our work includes:

(1) We propose a novel scheme that processes the 4D (dynamic 3D) volumetric data to extract multiple
    people's left and right hands and their associations with each individual in a  crowded
    and clutter scene.

(2) We propose a novel real-time people detection and tracking algorithm in 4D.
    We propose effective lightweight 2D features that enable fast people proposals and we propose new
    max-covering bipartite matching formulation for data association.
    To find people hands and their associations, we propose a new decomposition
    method to reduce a 3D regression problem to a sequence of 2D problems.

(3) Our algorithm gives strong performance comparing to different competing methods.
    Our method is not only much faster but also is more accurate, with nearly half the error of different competing
    methods.

\vspace{-5pt}
\subsection{Related Works}
\vspace{-5pt}

There have been many works on finding people's hands \cite{hand-track, hand-track2, hand-detection} and key points
in 2D color images \cite{deva,openpose} and in depth images \cite{kinect}.
These 2D estimations can be augmented to 3D joint locations by using triangulation \cite{cmudome, total, pose-multiview2}
and structure optimization \cite{pose-multiview1}
or using the depth information from the depth images \cite{kinect}.
With deep learning, 3D human pose can also be estimated directly from the 2D color images
\cite{single-pose, densepose}.
The 3D triangulation approach needs at least two
camera views to localize a body key point. When few cameras are used, due to object self- and inter-occlusions,
triangulation
becomes more and more unreliable as the clutter and number of people increase.
The direct 2D-to-3D approaches using color or depth images also have difficulty when dealing with self- and inter-occlusions
among people and objects
since each camera can only see one view of a person. For crowded and cluttered scene, we prefer to
use multiple sensors and methods that can take advantage of the multiple views.
In the paper, we propose a method that is able to process the dynamic volume which integrates all the
camera views
for accurate people detection and
hand localization in real time.

There
have been few previous works that directly process the dynamic 3D volumetric data to detect people key points.
Most previous methods find key points in single view images and then aggregate the result.
The result aggregation is non-trivial and often requires
expensive optimization \cite{cmudome, pose-multiview1, pose-multiview2, total, mrgbd}
especially when the estimation from different views are
inconsistent.
Finding medial axis is a popular lightweight scheme to find the skeleton and key points of a person
in a clutter-free volumetric data \cite{bodygraph}.
A Laplacian eigenspace method \cite{manifold} has also been proposed
to extract a single person's body parts from a clean 3D volume.
These methods find difficulties in cluttered scenes.
In \cite{thesis}, a deep learning approach is proposed to detect person key points and pose.
This method also requires a segmented person point cloud as the input.
The clean segmentation requirement limits the usability of these previous approaches in real applications,
because an accurate segmentation of each subject from clutter is a difficult task.
Different from these previous volumetric methods, our approach gives reliable results even in cluttered and crowded scenes.

Hand point localization can be tackled as a deep regression problem: a person volume is sent to a deep neural
network which directly outputs the coordinates of the hands of the person centered in the volume.
Such a method is an extension from DeepPose \cite{deeppose} on 2D color images.
With proper design, 3D deep regression gives decent results,
but it often fails when people poses are drastically different from the training
examples.

Our method is also related to the point cloud or volumetric data semantic segmentation
\cite{sparsenet,pointnet,synspeccnn,segcloud,octtree1,octtree2,kdtree}.
3D semantic segmentation labels points or voxels in a volume with specific classes.
Potentially it can be used to find the left and right hand voxels in the volume data
and then the locations of the left and right hand of the person can be further extracted.
However, 3D semantic segmentation methods are not sufficient for
hand detection and association because they are instance agnostic.
When there are multiple people
in a close proximity, we in fact want the hand labeling specific to each single subject.
Traditional 3D semantic segmentation methods thus have difficulty when associating the hands with subjects in a scene.
These methods also have high complexity even by taking advantage of the sparseness of the volumetric data.
Our proposed method solves the problem.


\vspace{-5pt}
\section{Method}
\vspace{-5pt}


The input of our algorithm is a fused point cloud from multiple RGBD cameras.
The point cloud gives a semi-complete scan of a dynamic environment that includes
people and different other objects. We voxelize the point cloud into a 3D occupancy
map, whose voxel size is $20mm$ in the $x$, $y$ and $z$ directions.
Examples of the volumetric data are shown in Fig.~\ref{fig-teaser}.
Directly finding people's left and right hands and their associations to each subject in the volume
is a challenging problem. Our proposed algorithm solves the problem efficiently.
Our method first finds the 3D bounding box of each subject and then finds
the hands associated with each subject centered in a 3D bounding box.
The details are as follows.

\vspace{-5pt}
\subsection{People detection}
\vspace{-5pt}
Let $V^t_{x,y,z}$ be the volume with spatial coordinates $x,y,z$ and time $t$.
$V$ represents
the 3D occupancy map, whose element is $1$ if the corresponding voxel is occupied and $0$ otherwise. For most
depth sensors, $V$ only represents the surface occupancy of different objects, i.e., the internal
part of different objects will have 0 occupancy just like the outside of the objects.
Even though we can also generate solid modeling of a scene, we found it does not
have much advantage in terms of performance compared to the simple surface representation.
In this paper,
we assume that $V$ only includes the surface voxels. 
The volume $V$ is rotated such that the floor plane is parallel to the $xy$ plane.

Quickly detecting a small set of people candidate proposals is a key step to achieve real-time hand detection and association.
Finding 3D person proposals is non-trivial.
The simple
method based on connected components is not sufficient to correctly find each person instance
in a cluttered and crowded point cloud.
Finding people in the 3D volume data using a 3D bounding box detector is one feasible solution.
However, brute force method has high complexity
due to the large amount of data in the volume representation.

\begin{figure}[tb]
\centering
\includegraphics[width=0.35\linewidth]{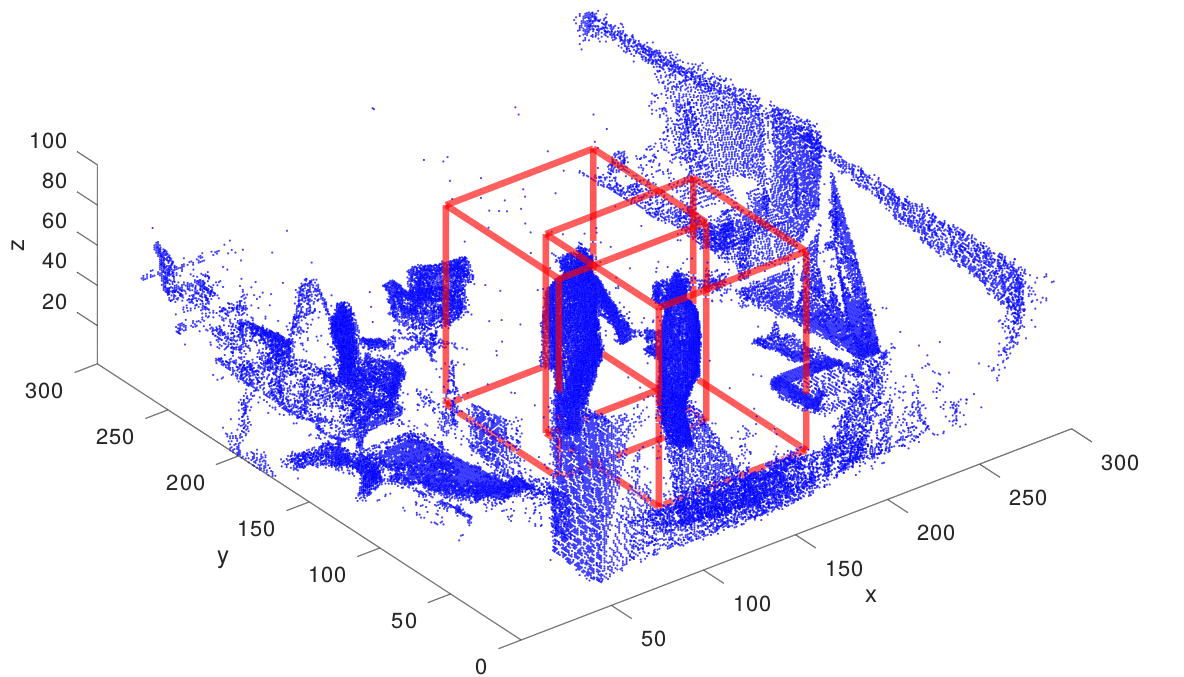}%
\includegraphics[width=0.216\linewidth]{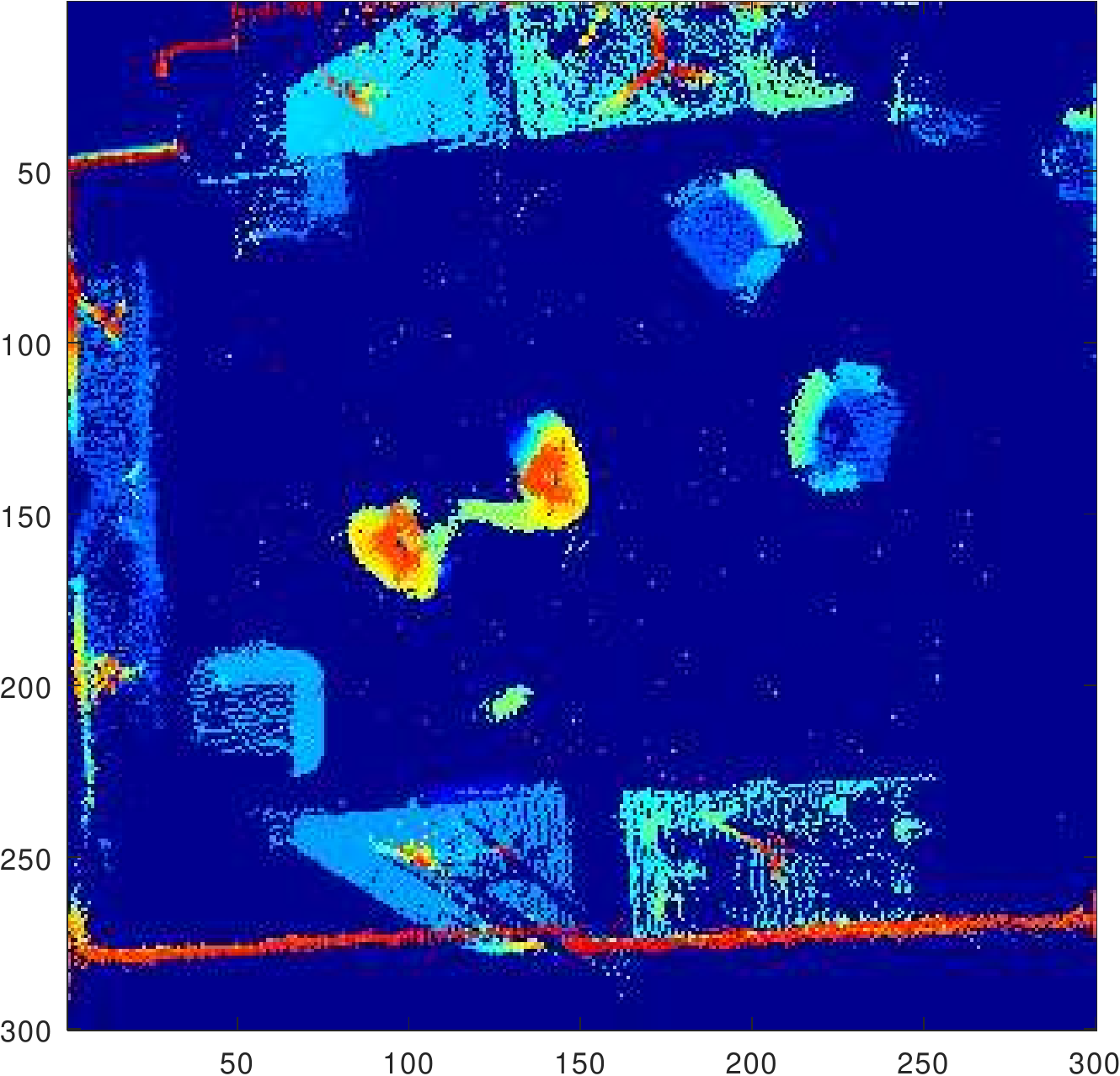}%
\includegraphics[width=0.216\linewidth]{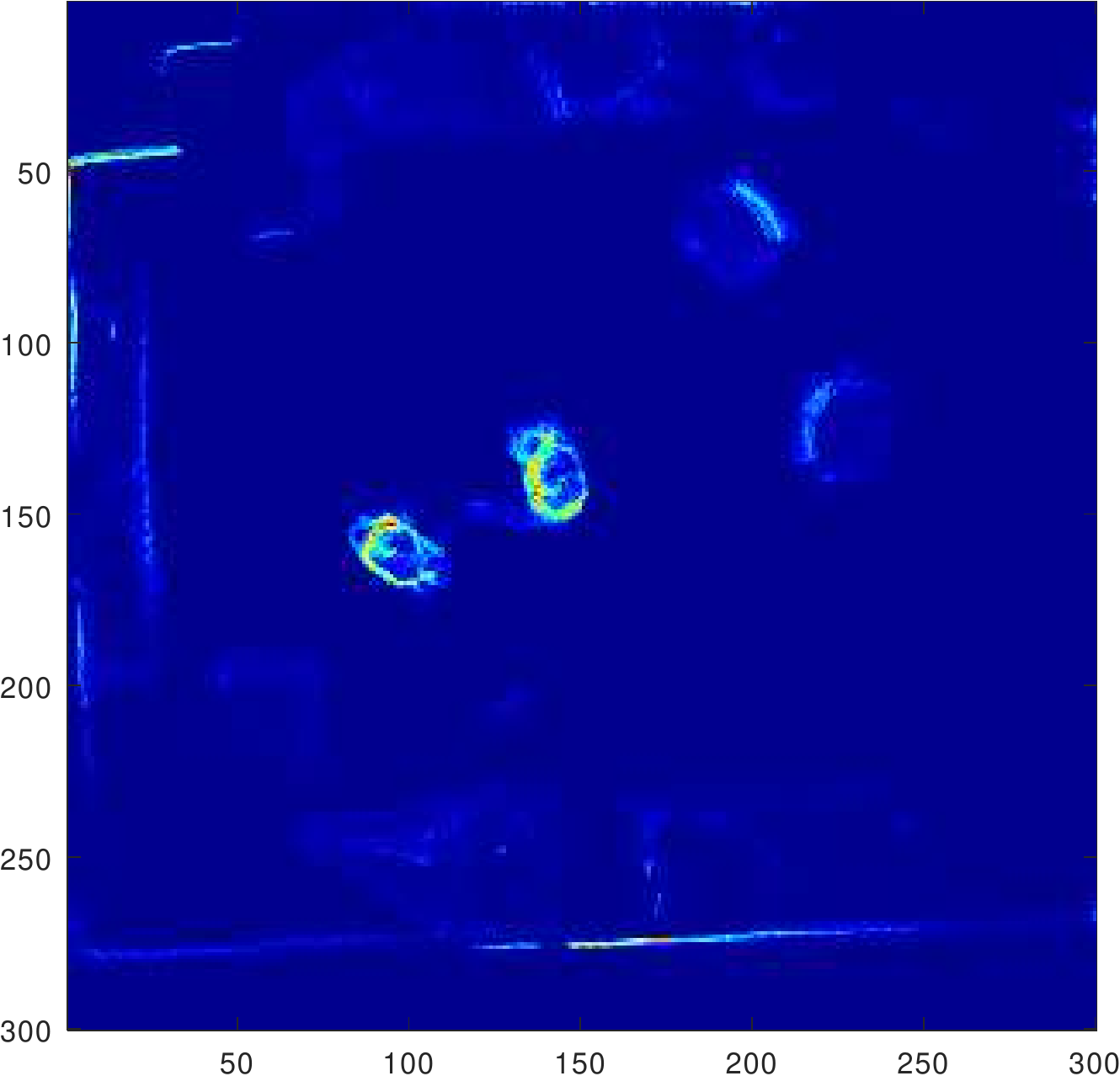}%
\includegraphics[width=0.216\linewidth]{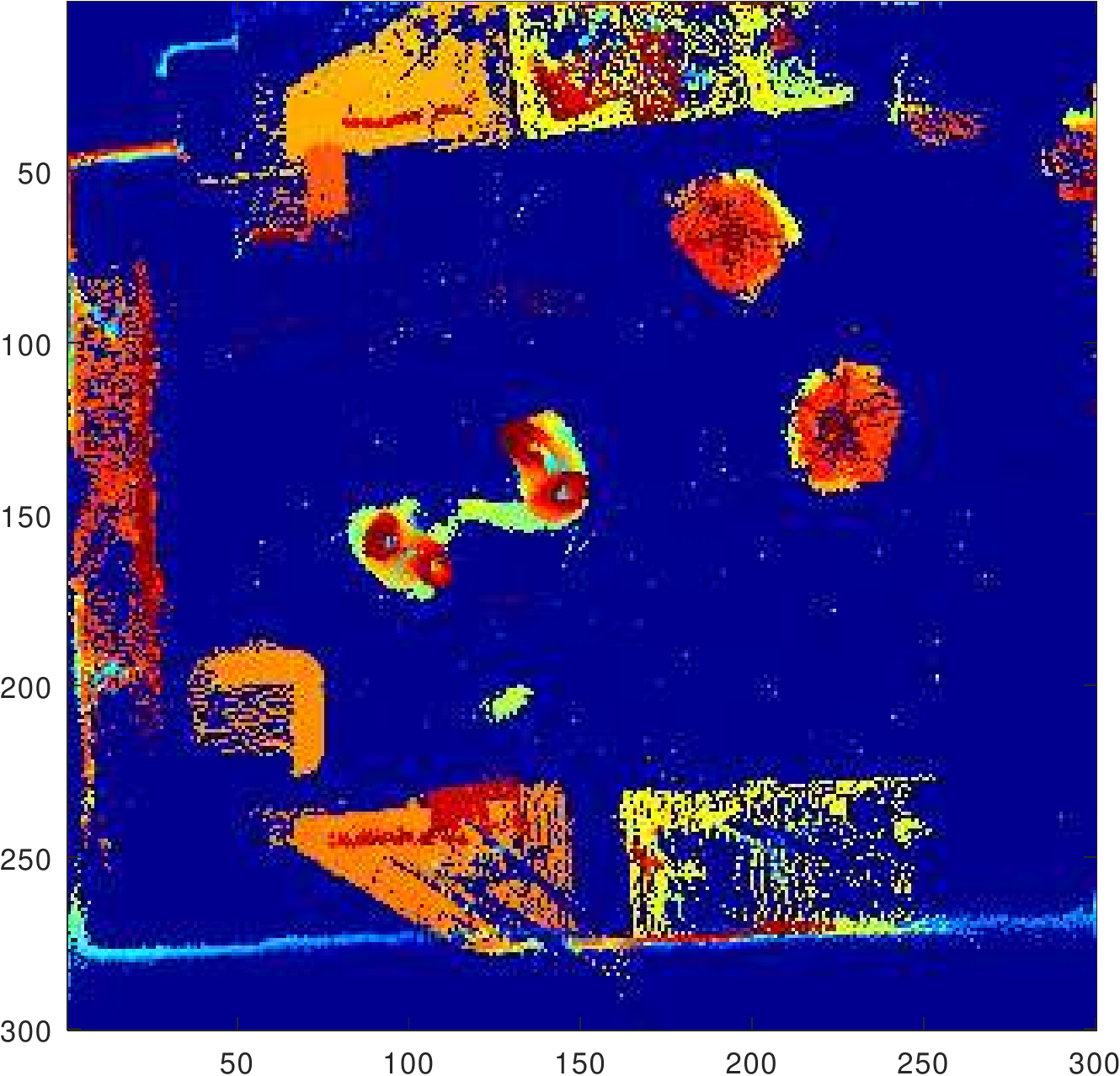}%
\caption{From left to right: the volume data, $f_t$ image, $f_s$ image and $f_b$ image.
$f_t$ and $f_b$ give the top-down and bottom-up view respectively. $f_s$ gives the voxel accumulated
value along each column. The red boxes in the volume are the people bounding boxes detected using the proposed method.
}
\label{fig-view}
\end{figure}

We propose an accurate lightweight solution as follows.
One key
observation is that people usually are apart in a top-down view as shown in Fig.~\ref{fig-view}. This suggests that
it is cost effective to find people candidates in such a view. Note that there is no requirement that
the cameras are strictly top-down. We can generate the virtual view from the volume. The top-down view image is
$f_t=\max_z(V*Z)$, where $Z$ is the mesh grid of subscript $z$ and $*$ is the element-wise product.
One top-down view of the volume is illustrated
in Fig.~\ref{fig-view}. The top-down view map $f_t$ clearly shows people and different objects in the space.

\begin{figure}[tb]
\centering
\includegraphics[width=0.5\linewidth]{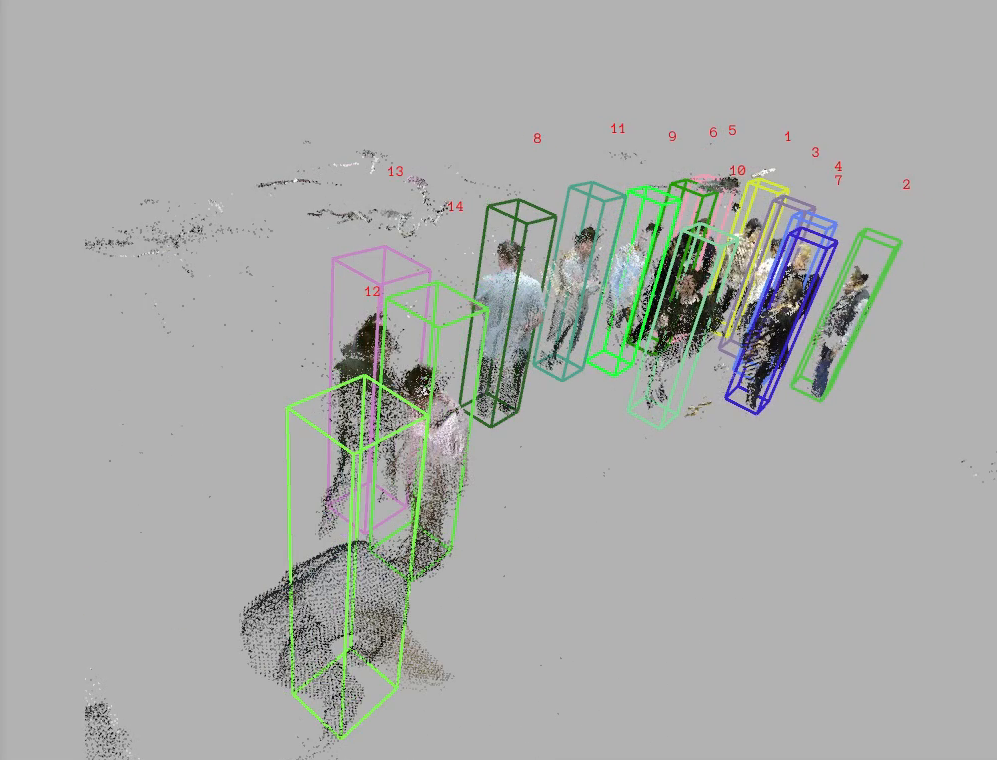}%
\includegraphics[width=0.5\linewidth]{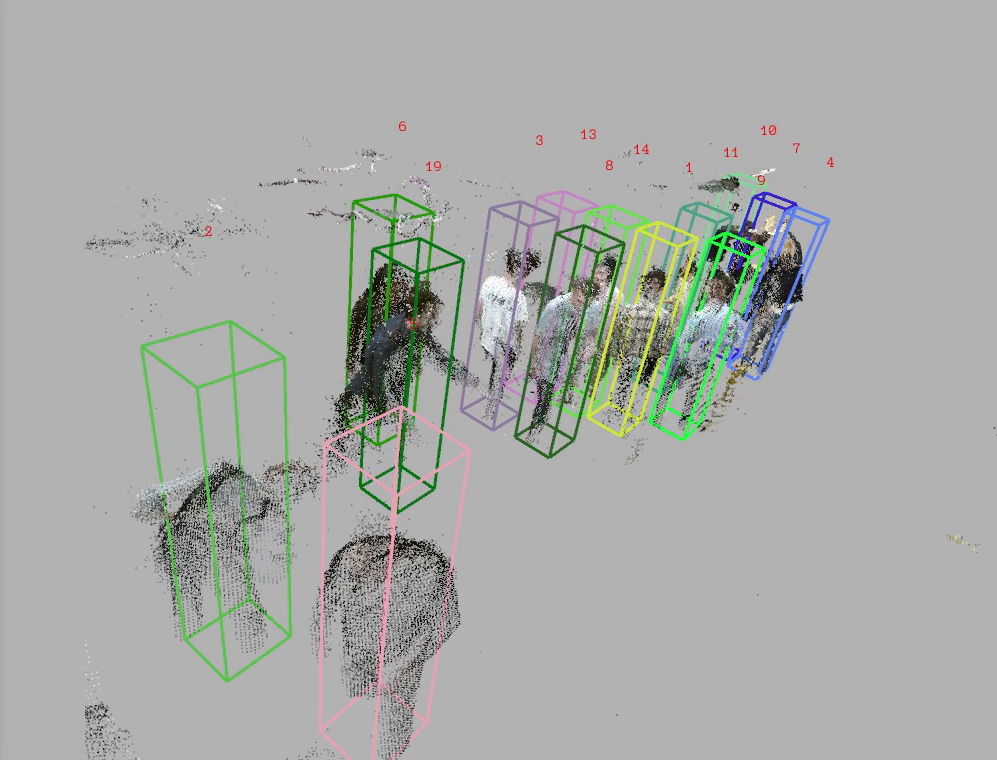}%
\caption{People detection and tracking in 4D using the proposed method in an environment with twelve Kinect-v2 cameras.
}
\label{fig-track12}
\end{figure}

Finding the people proposals only from $f_t$ is not sufficient.
One difficulty of the top-down
view image is that big object on top of a person may occlude the person below and affect the people detection.
To solve the problem, we further compute the summation of the voxels along the $z$ axis $f_s=\sum_z{V}$, as shown in
Fig.~\ref{fig-view}.
Now the side of the person is clearly identified even when the head of the person is occluded.

To further reduce the overhead occlusion problem in the top-down view representation, we also use
a bottom-up view of the volume, which is $f_b = \max_z{V*(N_z-Z)}$, where $N_z$ is the height of the volume $V$.
When computing $f_b$, we remove all the points on or blow the ground level.
An example of $f_b$ is shown in Fig.~\ref{fig-view}.
By normalizing $f_t$, $f_s$ and $f_b$ by $N_z$ and stacking them along the depth channel, we construct a color image
for people candidate detection. Note that our tracker does not need to keep $V$ for a large space, instead
it directly computes three $f$ maps. Later, the hand extraction module reconstructs the person sub-volume based on the
tracked person $xy$ locations. This can be done very efficiently and uses a small amount of memory.

We find people proposals using the three $f$ maps.
To achieve high efficiency,  our people
detector runs in a cascade fashion: first a linear classifier is trained using SVM to quickly reject a large
set of apparently non-people candidates and then a CNN is used to further verify the results.
Let the linear classifier weight be $W$ of size $51\times51$
and the bias $b$.
Then $W \otimes f_t + b$ can be used to classify the image as people
or non-people by its sign, where $\otimes$ is the convolution operator.
This is essentially a one-layer convolutional network.
As expected, $W$ has a bell shape which matches a person's shape from a top-down view. To increase the recall
of the classifier, we reduce the threshold from 0 to $\delta$ which is a small negative number.
The simple candidate proposer catches almost all the people candidates while at the same time has a quite
high false alarm rate. We further use a more expensive deep network to verify whether each candidate is a person or
not. The deep network works on the full $f$ image that has three channels $(f_t, f_s, f_b)$.
The network is a modified AlexNet \cite{alexnet}, with an input size of $51\times51\times3$ and output number of $2$.
The first convolution layer's stride is changed to 1 and its kernel to $3\times3$. The modified AlexNet
uses pretrained coefficients on ImageNet for all the layers except the first convolution layer and the last fully connected layer.
The network has small
input size and simple network structure. It is efficient in training and inference. The people detector
achieves above $98\%$ accuracy and recall.

\begin{figure}[tb]
\centering
\subfigure[]{\includegraphics[width=0.1\linewidth]{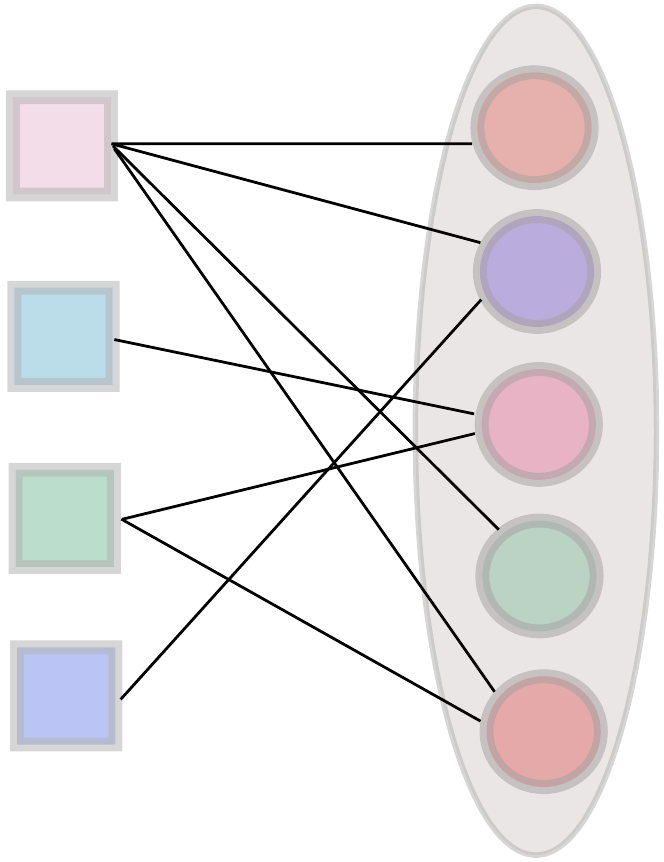}}%
\hspace{5pt}
\subfigure[]{\includegraphics[width=0.125\linewidth]{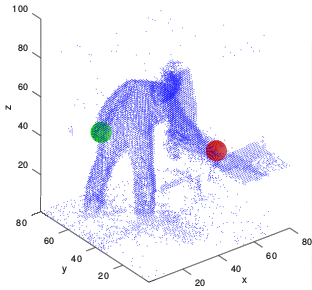}}%
\hspace{2pt}
\subfigure[]{\includegraphics[width=0.125\linewidth]{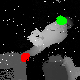}}%
\subfigure[]{\includegraphics[width=0.125\linewidth]{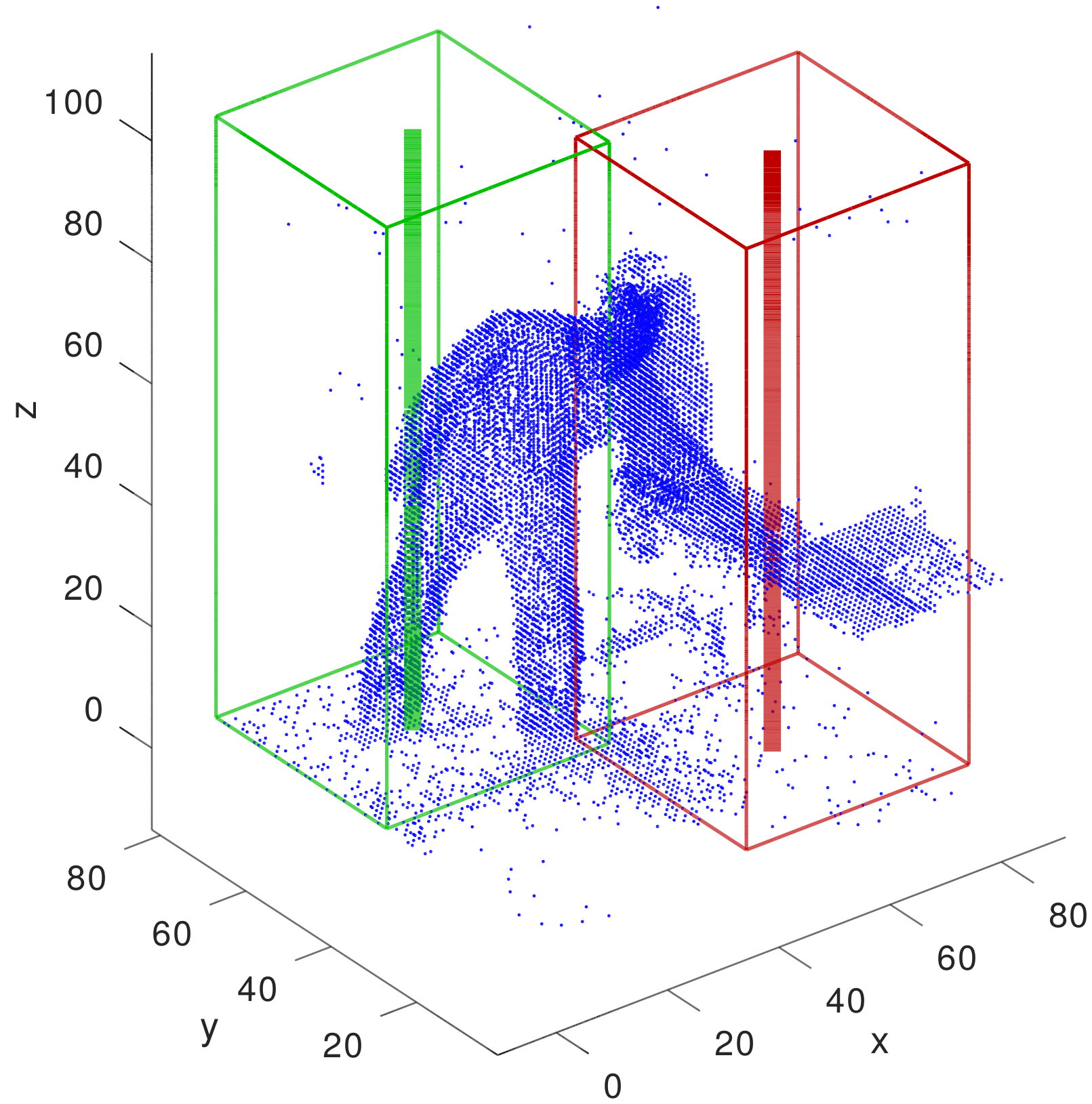}}%
\hspace{5pt}
\subfigure[]{\includegraphics[width=0.15\linewidth]{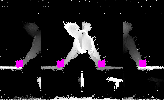}}%
\hspace{2pt}
\subfigure[]{\includegraphics[width=0.15\linewidth]{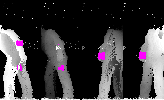}}%
\vspace{-4pt}
\caption{(a): Max-covering bipartite matching. The rectangle nodes are the trajectory nodes.
The round nodes are the people candidate nodes. The large oval indicates the max-covering constraints.
(b): Hand detection of our method.
(c): The top-down hand localization FCN result.
(d): Two thin volumes (red and green) for each hand based on the top-down FCN result.
(e): FCN result for the 4 side views of the left hand volume.
(f): FCN result for the 4 side views of the right hand volume.}
\label{fig-max-dec}
\vspace{-10pt}
\end{figure}

\vspace{-5pt}
\subsection{Max-covering bipartite matching}
\vspace{-5pt}
The fast person detector is accurate but still may miss detecting people or generate false alarms.
To solve the problem, we
track each subject in the volume to remove the false alarms and fill up the gap for missing detections.
Each person proposal is an $80\times80\times100$ volume with the bottom centered at voxel $(x,y,1)$, where $(x,y)$
is from the person proposer.
We maintain the trajectories of a set of tracked people.
Each trajectory keeps the tracked subject's appearance model (average color histogram),
motion vectors and
the volume centers in the past $N$ frames, where the max $N$ is 1000, and a person confidence score that
is the running average of person probability from the AlexNet people detector over all the tracked frames.
The trajectories grow as the volume input comes in.
A trajectory extends its length
and
updates all the states
if it finds a matched person candidate in the current frame.
If a trajectory has no matched person, the trajectory extends to a predicted position using motion estimation
without updating the appearance model.
The motion estimation uses the fast Lucas-Kanade method \cite{lucas} on the $f_t$ image with the key point centered
on each person proposal, which is either from the people detection or from a previous motion prediction.
At the predicted location, the people classifier updates the person confidence score.
For any current person candidate which is not matched by a trajectory, a new trajectory starts.
A trajectory is removed if the aggregated person score is lower than a threshold.

Trajectory extension is a matching problem as shown in Fig.~\ref{fig-max-dec}(a).
Apart from short distance and appearance similarity, we also want the trajectories to
cover as many people as possible in the scene.
We therefore
find the matching $X$ that minimizes the
energy, $E(X) = D(X) + A(X) - P(Y)$, with the constraint that the numbers of 1s in the permutation matrix $X$
equals the min of the rows and columns of $X$.
$Y$ is the margin of $X$ along each
column and thus represents the candidates chosen in the matching,
and $P(Y)$ is proportional to the total person probability of all the matched people candidates.
$D(X)$ is proportional
to the total distance between the last positions of the trajectories to the matched current frame candidates. $A(X)$ is the
Euclidean distance between color histograms of the trajectory appearance models and the current frame candidates.
The energy function minimizes the matching costs $D$ and $A$ and maximizes the people coverage $P$.
The max people coverage formulation is critical to achieve reliable tracking for very complex scenes.
The max-covering matching problem can be converted to a generic bipartite matching problem and solved using
primal-dual algorithm,
since the energy of $P$
can be distributed to different edges of the bipartite graph similar to the terms of $D(.)$ and $A(.)$.
It is easy to show that in such a setting optimizing the bipartite matching is equivalent to minimizing
the max-cover matching energy.
The matching graph is very sparse due to the fact that a person cannot travel very far in a small time interval.
Each trajectory has at most 10 connections to the people candidate nodes.
This enables fast matching whose complexity is proportional to the number of candidates, comparing to the cubic complexity of
dense matching.
Fig.~\ref{fig-track12} shows our people detection and tracking results using 12 Kinect-v2 cameras.

\vspace{-5pt}
\subsection{Hand localization and association}
\vspace{-5pt}
With each subject detected and tracked, we cut out person sub-volumes and further find the left hand and right hand of the person
centered in each sub-volume. In this paper, the sub-volume has a size of $80\times80\times100$ voxels, and each voxel has
edge length of 20mm.
The loose bounding box of each target often also includes other people or objects in the proximity.
The person may have different body poses, orientation and the data is often noisy.
Our method needs to be robust against all these factors.


\begin{figure}[tb]
\centering
\includegraphics[width=0.6\linewidth]{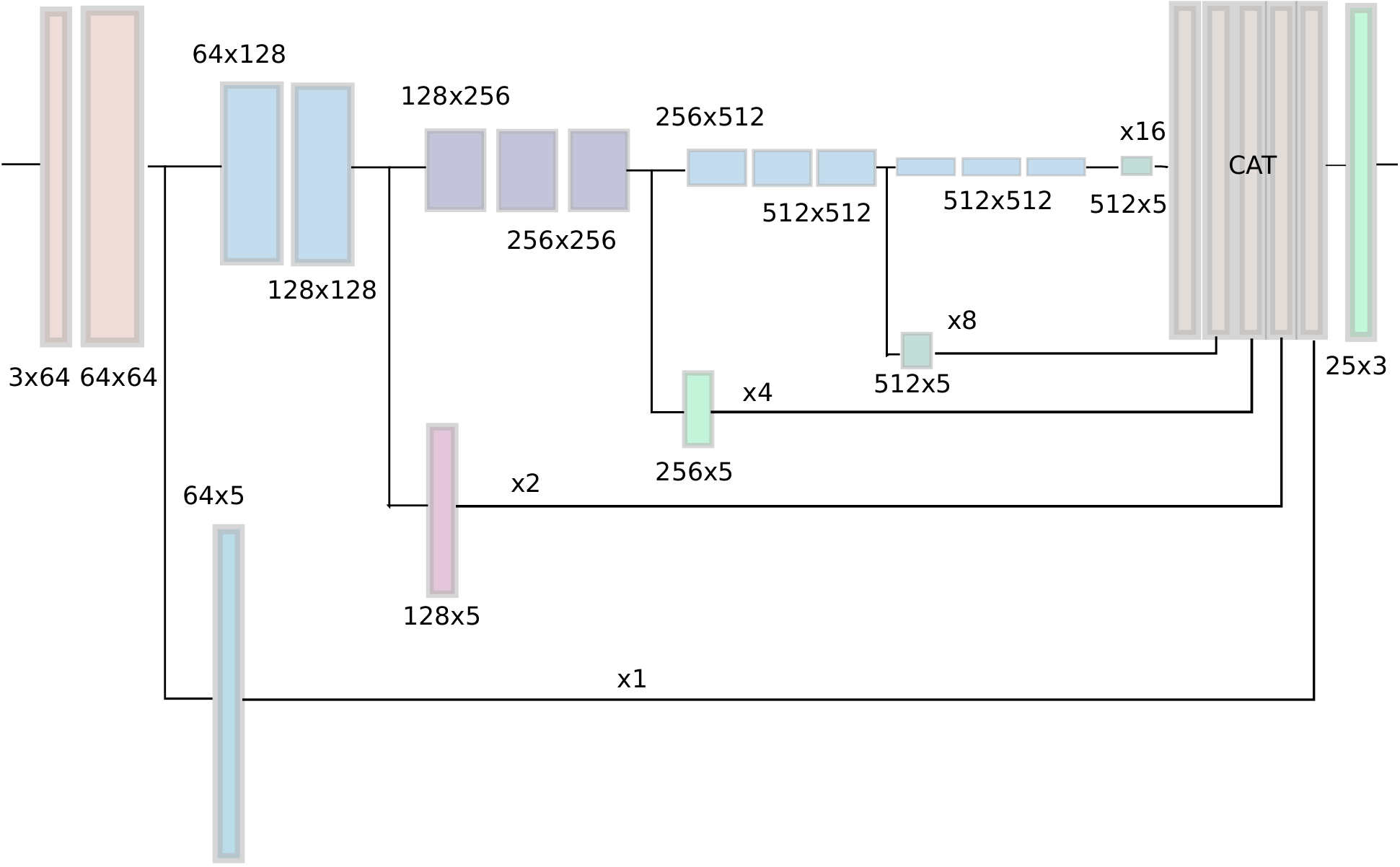}%
\caption{FCN for top-down and side-view semantic segmentation.}
\label{fig-fcns}
\vspace{-10pt}
\end{figure}

Directly processing the 3D volumetric data would be quite slow.
We propose a decomposition approach, which is much faster and at the same time
it gives much better results than the deep regression or the 3D semantic segmentation approach.
The basic idea is that we can decompose the 3D key point localization
into a sequence of 2D problems: we first determine the $(x,y)$ locations of the hands from the top-down view of the volume and
then we cut out two thin volumes centered at the left and right hand locations. From the eight side views
 of the cropped
thin volumes we estimate the $z$ coordinates of the two hands.
To determine the key points in each of the above nine views,
we design deep convolutional neural network for semantic segmentation.
Fig.~\ref{fig-max-dec}(b-f) give an example of converting 3D localization to inferences on a sequence 2D images.

In the following, we give the details of the proposed method. Centered on each person location, we extract an image patch
of $80\times80$ pixels
on image $f_t$ to localize the left and right hand on the $xy$ plane.
We train a fully convolutional network (FCN) to classify each pixel as
left hand, right hand or no hand. The network structure is shown in Fig.~\ref{fig-fcns}.
The FCN uses 13 convolution layers of VGG16 \cite{vgg16} as the backbone and is
a variant of the HED network \cite{hed} which has been used for edge detection.
Skip layers are introduced to generate deep features at
different scales. The deep features are concatenated at each pixel location and passed through a classification network for
semantic segmentation. The VGG16 backbone is pretrained on ImageNet.

Note that a fully convolutional network's translation invariance is an undesirable feature in our application, because
we would like to classify the hands of the center person in the $80\times80$ image patch as the positive labels, while
all the other pixels including the hands of surrounding people should receive negative labels. Interestingly, our FCN
in fact does a great job to distinguish the person of interest at the center of the image patch
even when the distracting person is very close. The reason this works is due to the small size of the input image patch
and
the padding when we do the convolutions. There are four stages of max-pooling and thus the effective zero padding is 1, 2, 4, 8, and 16 pixels
for the five stage of convolutions and the effective kernel sizes go from 3 to 48. The big black boundary after padding helps
give different results near the boundaries from the center regions of the input image when convolution is applied.
An seeming artifact gives an interesting desirable result, without the need to introduce position guidance map into
the input.


The hand FCN of the $xy$ plane is accurate. To further extract the $xy$ coordinates of the left and right hand centers,
we simply compute the median of the coordinates of the pixels predicted as the left and right hand.
After predicting the $x$ and $y$ coordinates of the hands, we further estimate the corresponding $z$ coordinate.
One possible way is to use the side views of the person volume, similar to the multiple view 3D object classification and segmentation method
\cite{2d3d2, 2dfor3d}.
The difficulty of this approach is that, when there is a crowd, it is likely the person at the center is occluded from all the side views.
To solve the problem, instead of using the person side views we cut two thin volumes with the centers at the left and right
hand position estimated from the top-down FCN. In this paper, the thin volumes have a dimension of $41\times41\times100$.
One example is shown about how such slicing is performed in Fig.~\ref{fig-max-dec}(d).
The side FCN, with similar structure as the top-down FCN, is applied to all the eight side view images for two hands.
The side FCN has the same structure as the top-down FCN except that it has two output channels instead of three.
The final $z$ coordinate of a hand is the median of the $z$ coordinates from the hand's four side view images.
Fig.~\ref{fig-max-dec}(c, e, f) show the top-down view and thin volume side view FCN classification results
and Fig.~\ref{fig-max-dec}(b) shows the hand localization result.
Due to the redundancy of the vertical slices
the method is robust against FCN classification errors.

It seems counterintuitive that a sequential 2D approach can give better results
than a full 3D convolutional network since the hands are mostly occluded if a person has a standing pose.
So the proposed 2D method has to hallucinate the left and right hand estimations on the top-down image. This in fact is not
a weakness since we found that the most challenging case to find hands
for the 3D convolutional network is when a person is standing
still and his/her hands merge with the legs. In such a case, finding the hands from top-down views
is as effective as the full 3D view by using shoulder as a context.
From the training perspective, the decomposition approach is also helpful.
By using decomposition,
we inject more human heuristics and this in fact helps a simpler
network to achieve much better results.

\section{Experiments}
\vspace{-5pt}

We evaluate the proposed method on data with ground truth for quantitative comparison with competing methods
and the data without ground truth for qualitative analysis.

\vspace{-5pt}
\subsection{Experimental setup}
\vspace{-5pt}
We collect a dataset of 20 subjects in a 300 square feet lab space covered by 7 Kinect-v2 cameras. The Kinect-v2 cameras are mounted
near the ceiling with a roughly 45 degree downward angle.
The Kinect-v2 depth images which are aligned to the color images are downsampled to $320\times160$ and
streamed through a gigabit network using UDP multicast.
The images are received at the center computer and processed in real time.
There is no guarantee that the received images are completely synchronized. Our method has to deal with
all the data defects.

To extract the ground truth hand points automatically, the subject of interest in each recording session wears
a green glove on the right hand and a red glove on the left hand. Each recording session includes one person of
interest and a few people acting as distracting subjects.
The people in the recording perform different daily activities and have many interactions.
The scene recorded also includes non-people objects such as chairs, tables, plates, cups, and closets.

With simple color segmentation, color gloves help generate accurate ground truth of the hand semantic segmentation
in 3D volumes and we further obtain the 3D coordinates by computing the median of the $xyz$ coordinates for each hand.
Since our method does not use color in training and testing, the color glove does not affect
the depth data and the method evaluation.

The 4D people detection is trained using around $500k$ volumes from a different lab that includes 12 Kinect-v2 cameras.
Our people detection and tracking achieves $100\%$ accuracy on the hand localization training and test videos.
After cropping out the person of interest volumes and
manual data cleanup,
we have a total of 42856 3D person volumes
of size $80\times80\times100$ which has a voxel size of 20mm.
Among the 20 people, we randomly choose 15 people's data (30481 person volumes) for training and the rest of the 5 people
(12375 volumes) for testing.
We do not leave out validation data and we will explain how to
make a fair comparison among different methods in such a setting.

\begin{figure*}[tb]
\centering
\includegraphics[width=\linewidth]{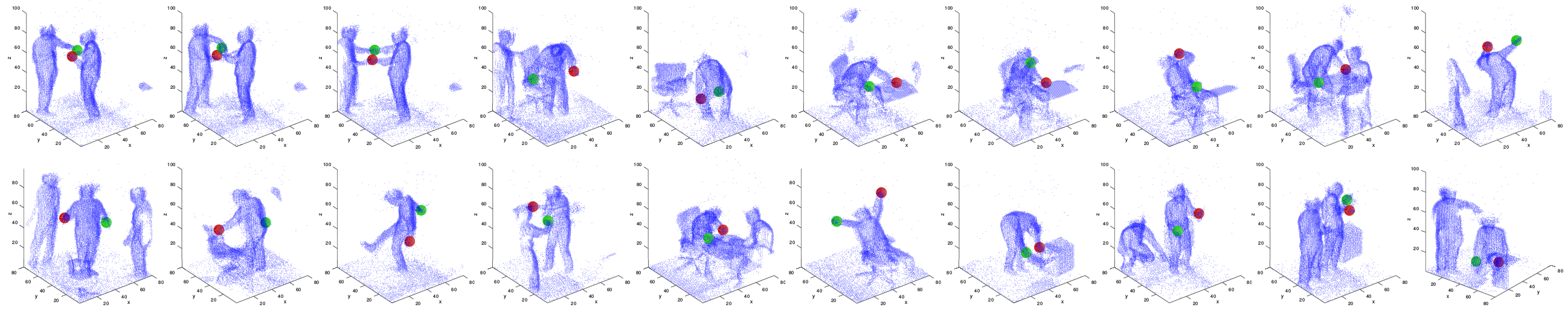}%
\linebreak
\includegraphics[width=\linewidth]{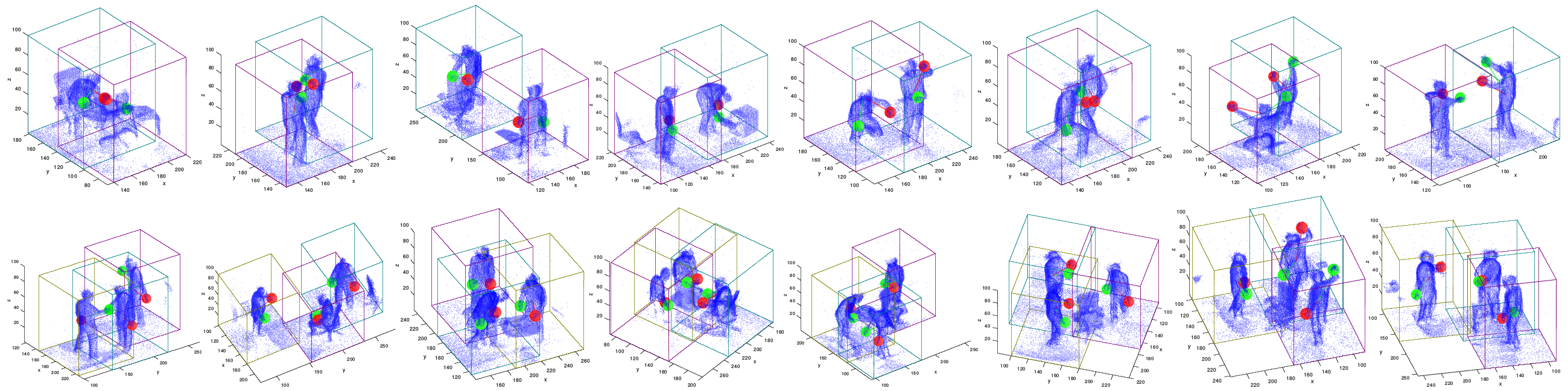}%
\caption{Rows 1-2: Sample results of our method on the ground truth test data of five subjects,
each of which is centered at a tracked volume.
Rows 3-4: Sample results on every subject in the scene.
The green dot indicates position of the right hand and the red dot the left hand. The result is overlaid on the volume data
from the tracking result, with the target person located at the center of each 3D bounding box.
}
\label{fig-gtsamples}
\vspace{-5pt}
\end{figure*}

\begin{figure}[tb]
\centering
\includegraphics[width=0.75\linewidth]{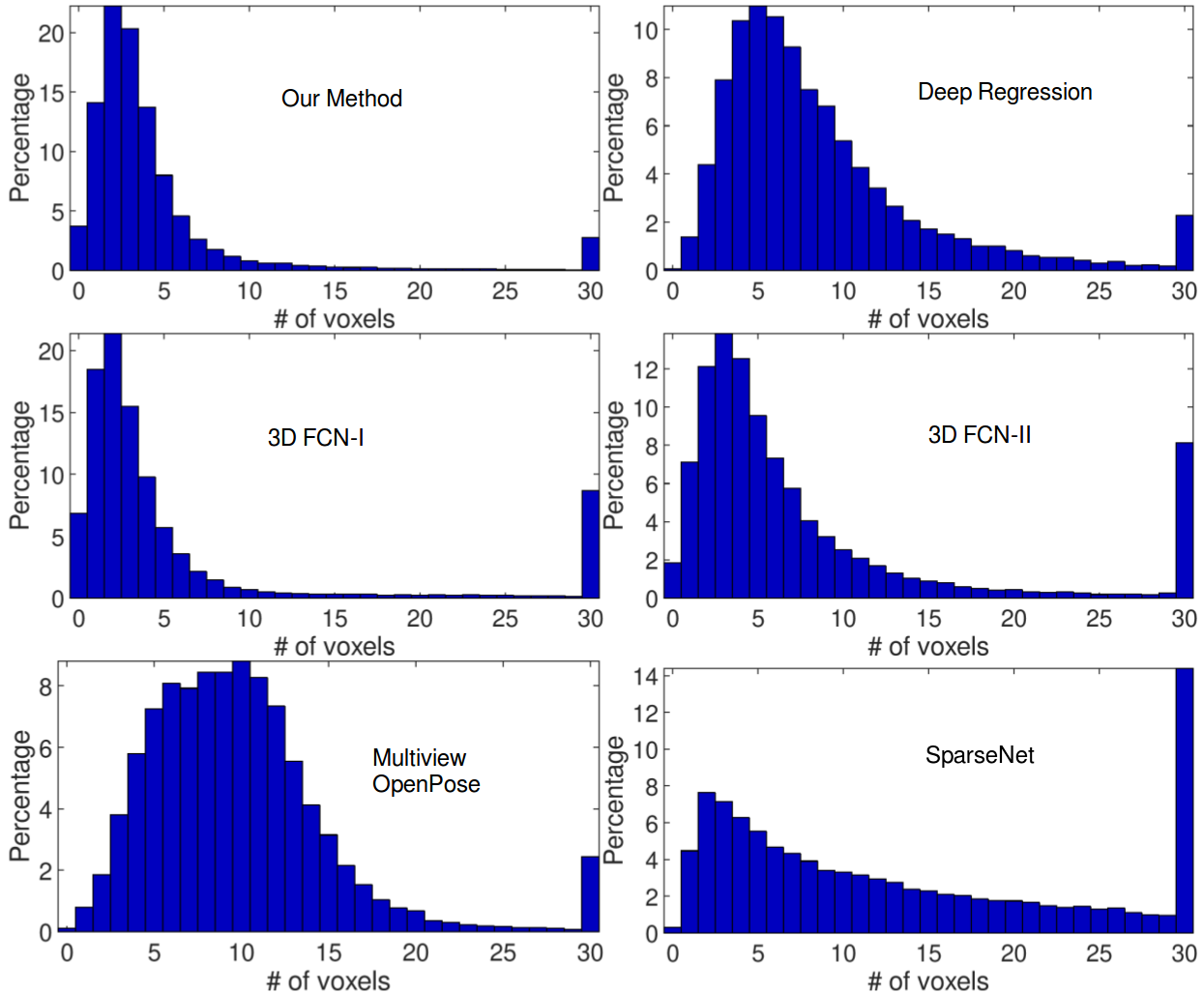}%
\caption{The histograms of the hand localization errors of different methods.
The size of the voxel is $20\times20\times20mm$. 3D FCN-I and II use the weights of 10 and 100 respectively in the
weighted cross entropy loss. }
\label{fig-hist}
\end{figure}

\vspace{-5pt}
\subsection{Competing methods for comparison}
\vspace{-5pt}
There are no available methods that can be directly used to localize hands in the cluttered 4D point clouds
that involve multiple closely interacted people.
We adapt several 3D deep learning methods to our application. All the volume based competing methods use our
tracked volumes.
Kinect SDK provides 3D skeleton and hand point localization; but since those methods are designed for games, they
work poorly for our camera angles and cluttered scenes. We do not include them in the comparison.
We use multiple view OpenPose \cite{openpose} method instead.
The OpenPose approach also depends on our tracker to separate multiple people
in the scene.
The competing methods are as follows.

\textbf{Deep regression}: We construct a deep regression network with
3D volume data as the input and it directly outputs the normalized $xyz$ coordinates of the two hand locations.
Here each coordinate is divided by the max value along that dimension so that its coordinate is normalized to
[0,1].
The regression network includes 4 convolution layers and 3 fully connected layers.
The kernel size for all the convolution layers is $3\times3\times3$. The numbers of output channels are 8, 8, 16 and
16 for the four convolution layers and the output channel numbers are 4096, 4096 and 6 for the
three fully connected layers. ReLU and 3D maxpooling are inserted after each the convolution layer.
The output of the fully connected network goes through
a Sigmoid layer to map the output to be in [0,1].
The structure of the network has been tweaked to achieve the best performance of hand regression.
During training, to augment the data,
the input volume and the output coordinates
are rotated by a random degree in [0, 360]. $50\%$-dropout is also applied before the
first two layers of the fully connected layers during training.

\begin{figure}[tb]
\centering
\includegraphics[width=0.75\linewidth]{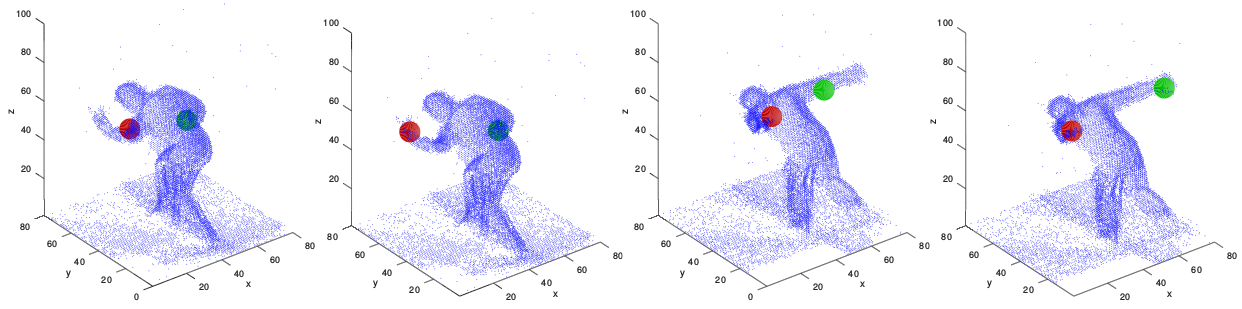}%
\caption{For human poses with large difference from the training data, deep regression (shown in the odd columns) fails to
localize hands,
while our methods give the correct result as shown in the even columns.
 }
\label{fig-reg}
\vspace{-10pt}
\end{figure}

\textbf{3D FCN approaches}: 3D fully convolutional network can be constructed to give the voxel
level classification. By finding hand voxels, we can localize the hands in the 3D volume.
We tested two kinds of networks. We construct one 3D fully convolutional network
and the other is the SparseNet \cite{sparsenet}.

Our 3D FCN has the input
size of $80\times80\times100$. It has 4 convolution layers followed by ReLU and 3D max-pooling.
The input volume is augmented into a 3-channel volume data.
The numbers of the output channels of
the 4 convolution layers are 16, 32, 64 and 64.
Each output from the convolution and ReLU layers are sent to
a 3D convolution layer with kernel size $3\times3\times3$ and 5 output channels, which are further
trilinearly upsampled to size $5\times80\times80\times100$. These five channel features are stacked along the channel dimension
and sent through a 3D convolution with kernel size $3\times3\times3$ and three output channels
to give the final result.
The input single-channel volume is augmented into three volumes, each of which is an element-wise product
of the Gaussian blurred occupancy map and the $x$, $y$ and $z$ mesh grid. The Gaussian blur helps the classifier to
tolerate local deformations and the $x$, $y$, $z$ coordinates help the FCN be
more position sensitive so that it can ignore the surrounding persons in an input volume.
All the parameters and the input format have been tweaked to make sure the performance of the 3D FCN gives
the best performance.

Besides our dense implementation of the FCN network, we also compare the results of the proposed method with SparseNet \cite{sparsenet}.
3D SparseNet is one of the state-of-the-art methods that directly work on the points instead of the volumes.
We use the UNet included with the SparseNet package for semantic segmentation. UNet is a more complex structure
than our 3D FCN and thus has potential to give better results.
During training, both the 3D FCN and the 3D SparseNet augment the data by rotating the input volumes and label volumes
with random angles from 0 to 360 degrees.

\begin{figure}[tb]
\centering
\includegraphics[width=0.75\linewidth]{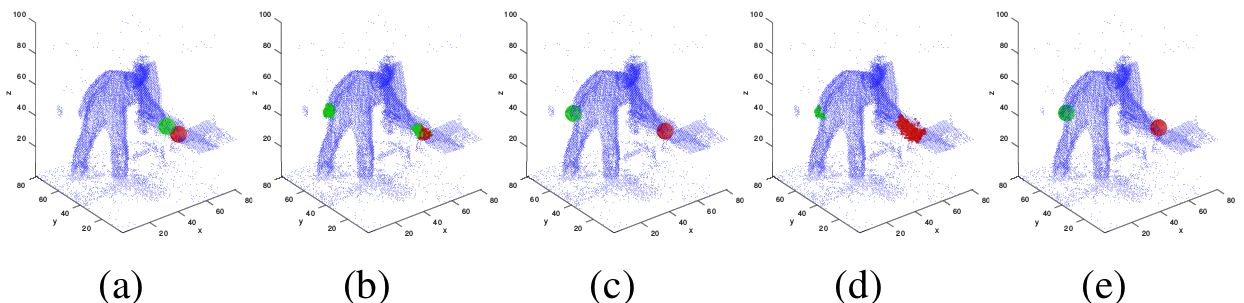}
\caption{Comparison with the 3D fully convolutional network (FCN).
In (a) and (b) are the 3D FCN-I hand localization and hand semantic segmentation result.
(c) and (d) show the FCN-II hand localization and hand semantic segmentation result.
FCN-I has weight 10 for positive samples in cross entropy loss and the weight for FCN-II is 100.
(e) shows hand localization of the proposed method.
}
\label{fig-fcn}
\vspace{-10pt}
\end{figure}

\begin{figure}[tb]
\centering
\includegraphics[width=0.75\linewidth]{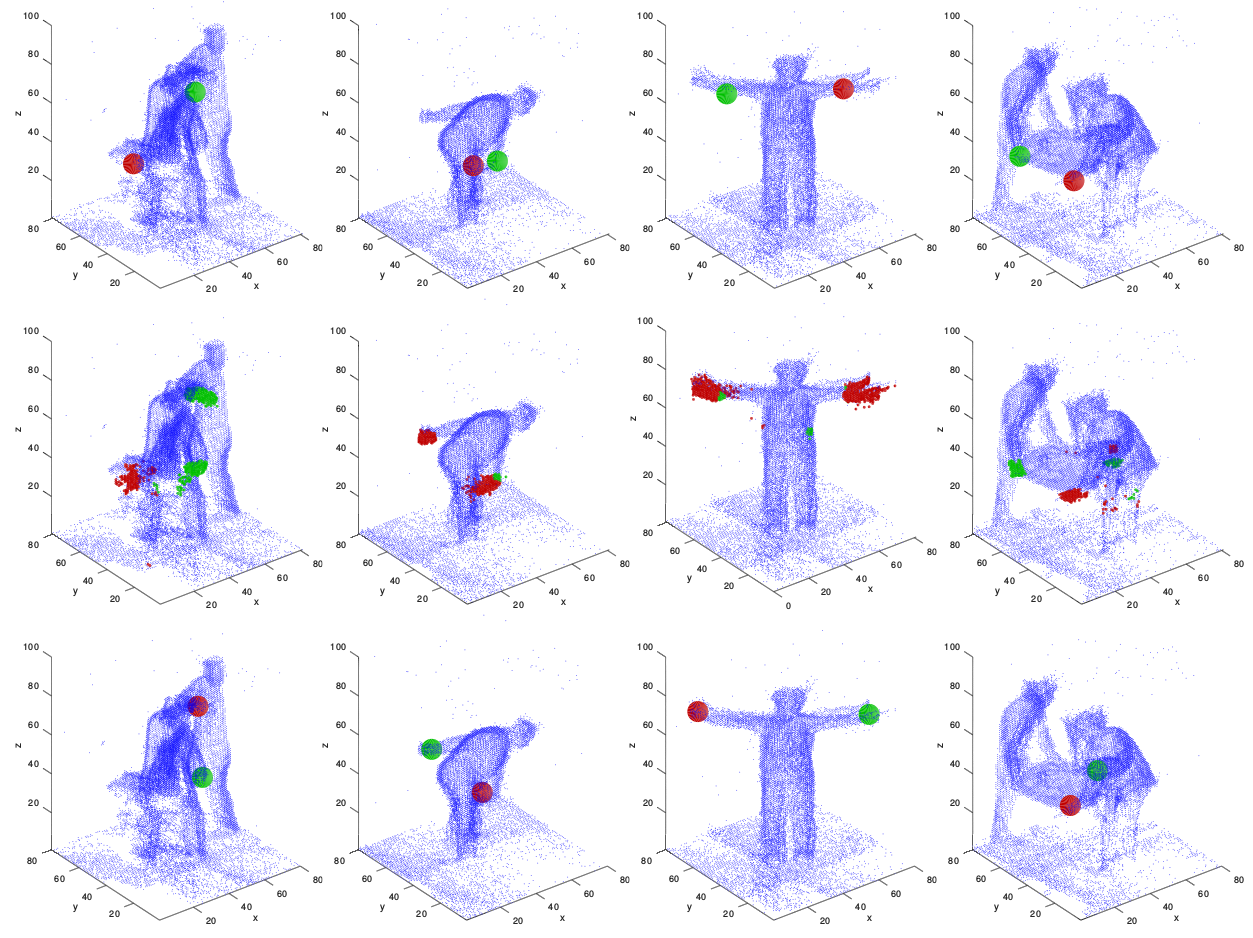}%
\caption{Row 1: SparseNet \cite{sparsenet} hand localization. Row 2: SparseNet semantic segmentation.
Row 3: Our result.
}
\label{fig-sparsenet}
\end{figure}

\begin{figure}[tb]
\centering
\includegraphics[width=0.75\linewidth]{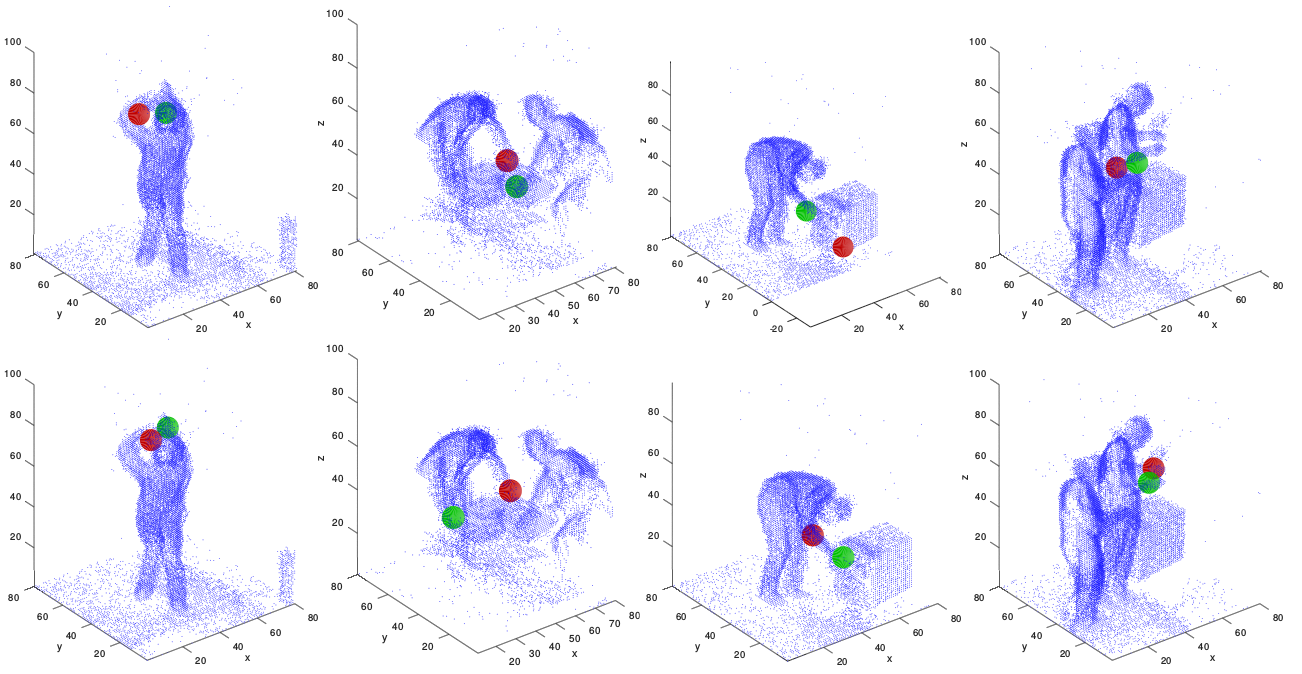}%
\linebreak
\includegraphics[width=0.75\linewidth]{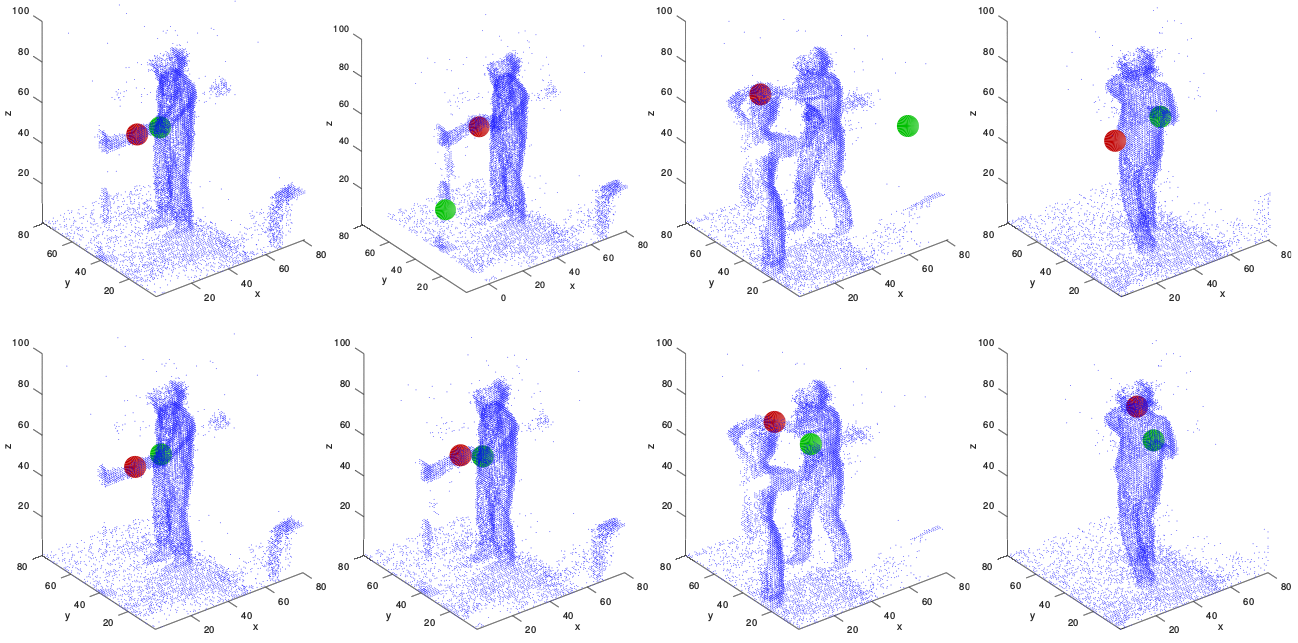}%
\caption{Rows 1,3: Sample results of multiple view OpenPose \cite{openpose} method. Rows 2,4:
Our hand localization result.}
\label{fig-openpose}
\end{figure}


\textbf{Multiple view approach}:
We also include a multiple view approach as a competitor. The multiple view approach uses
OpenPose \cite{openpose}, a state-of-the-art human pose and key point detector,
to extract 2D people joints in the RGB images of the seven views.
When there are multiple people in the scene, we need to correspond the key point estimations
of the same person in different views. This is achieved by projecting our 3D people bounding boxes to each view
and finding the skeleton whose bounding box has the maximal IOU with the projected bounding box.
Once the correspondence of the joints are determined, we triangulate the back projection rays to find the hand 3D coordinates.
Triangulation in 3D is a non-trivial task due to the 2D estimation errors. The optimal multiple 3D line ``intersections'' are also hard to
find. Min-square solution gives poor result even with one wrong back projection rays. To solve the problem, we use a robust
estimation method which projects each voxel in the whole person volume to each view and we find the voxel whose 2D projections are the closest
to the 2D key points in terms of the truncated $L_1$ norm. The truncation threshold is set to be 30 pixels.

\setlength{\tabcolsep}{5pt}
\begin{table}[tb]
    \centering
    \begin{tabular}{ c | c | c | c | c | c | c}
    \hline
                 & Our method & Deep-reg & 3D FCN-I & 3D FCN-II & SparseNet & MOpenPose \\ \hline \hline
    Mean      & 5.22       &  8.96    &  8.04        & 9.91         & 15.49         & 11.11 \\ \hline
    Std       & 9.63       &  7.89    &  15.87       & 15.34        & 16.07         & 14.52 \\
    \hline
    \end{tabular}
    \caption{Mean error and standard deviation of different methods. The unit is one voxel whose length is $20mm$.
            Manhattan distance is used to measure errors. Deep-reg is 3D deep regression. MOpenPose is the multiple view OpenPose \cite{openpose} method.
            }
    \label{tab-error}
\end{table}

\subsection{Comparison results}
In the ground truth comparison, our method takes the 30-epoch model for evaluation, while the methods of deep regression, 3D FCN, 3D SparseNet
all take the models that give the lowest average hand prediction errors on the test dataset among 100 epochs.
This gives clear advantage to the competing methods
in the comparison. The comparison with OpenPose is not strictly speaking fair because it does not use our training data but uses many more
other color images than our method. It also gives the wrist coordinates instead of the hand coordinates.
This may introduce a system error.
For the multiple view approach comparison, we pay more attention to the gross errors that are greater than 20 voxels.
During testing, if a hand is not detected, we assume its position is at the center of the person volume.

The sample results of the proposed method
are shown in Fig.~\ref{fig-gtsamples}.
The sample comparison results between our method and different competing methods
are shown in Fig.~\ref{fig-reg}, Fig.~\ref{fig-fcn}, Fig.~\ref{fig-sparsenet} and Fig.~\ref{fig-openpose}.
The mean errors and the standard
deviations are shown in Table~\ref{tab-error}.
The error distributions of different methods are shown in Fig.~\ref{fig-hist}.
The distance measure is Manhattan distance.

Our method gives the lowest mean error in the ground truth comparison.
The deep regression approach works reasonably well. However, due to the usage of a global model,
deep regression gives large errors when the pose
has large difference from any of the training samples.
3D FCN gives slightly lower mean error than the deep regression. But it gives more gross errors due to the imbalanced training data.
With a positive sample weight of 10 for the training of FCN-I, there are many missing hand detections.
After we increase the hand voxel weight from 10 to 100 in FCN-II's cross entropy loss, the mean error and the large gross errors both
increase as shown in Fig.~\ref{fig-hist}. And, as shown in Fig.~\ref{fig-fcn}, the hand classification accuracy suffers.
Adjusting the weight in the cross entropy loss is thus ineffective.
As shown in Fig.~\ref{fig-sparsenet}, SparseNet \cite{sparsenet} has a hard time distinguishing left and right hands of
the same person and the hands among different people in the close proximity.

As shown in Fig.~\ref{fig-hist}, our method also has fewer gross errors that are greater than 20 voxels than all the competing methods.
The multiple view OpenPose method gives many large errors. This is due to the smaller number of the cameras and there is a
high chance that only one camera can see a hand.

Fig~\ref{fig-gtsamples} Rows 3-4 show our results on every subject in sample test videos.
Recall that in each video only one subject who wears color gloves has the ground truth hand locations.
Our proposed method still gives good results, which shows that the color gloves play no role in the performance of
the proposed method.
Our method is also fast. Our person proposal and tracker method can handle
50 cameras and up to a few hundred people in real-time tracking using a single machine with a GTX1080TI GPU.
Our hands localization for each person volume
is around 5 milliseconds by using a GTX1080TI GPU,
while the dense 3D FCN takes 20 milliseconds and SparseNet UNet takes 30 milliseconds. The simple deep regression network's
inference time is around 5 milliseconds. Multiple view OpenPose approach needs around 700 milliseconds
to process the 7 RGB images from each view and more time
to optimize the 3D hand localization.

\section{Conclusions}
We propose a real-time multiple people hand localization method that directly works on 4D dynamic volumetric data.
Our method is robust against strong clutter, crowd, complex people-people and people-object interactions.
It gives more accurate result and at the same time is faster than different competing methods.
We believe the proposed method can be used to enable many applications such as human computer interaction,
gesture recognition, activity recognition and human object interaction understanding.

\end{document}